\documentclass[12pt,]{article}
\usepackage{authblk}
\usepackage{fullpage}
\usepackage{amssymb,amsmath}
\usepackage[utf8x]{inputenc}
\usepackage[T1]{fontenc}
\usepackage{siunitx}
\usepackage[version=3]{mhchem}

\usepackage[export]{adjustbox}

\usepackage[english]{babel}

\usepackage{url}
\usepackage{natbib}
\usepackage[left]{lineno}

\usepackage{setspace}
\usepackage[unicode=true]{hyperref}
\hypersetup{breaklinks=true,
            colorlinks=false,
            pdfborder={0 0 0}}
\usepackage{graphicx}
\graphicspath{{figures/}}
\usepackage{subcaption}
\usepackage{caption}
\makeatletter
\def\ScaleWidthIfNeeded{%
 \ifdim\Gin@nat@width>\linewidth
    \linewidth
  \else
    \Gin@nat@width
  \fi
}
\def\ScaleHeightIfNeeded{%
  \ifdim\Gin@nat@height>0.9\textheight
    0.9\textheight
  \else
    \Gin@nat@width
  \fi
}
\makeatother
\setkeys{Gin}{width=\ScaleWidthIfNeeded,height=\ScaleHeightIfNeeded,keepaspectratio}%

\title{Choosing an Appropriate Platform and Workflow for Processing Camera Trap Data using Artificial Intelligence}
\author[1]{Juliana V{\'e}lez}
\author[2] {Paula J. Castiblanco-Camacho}
\author[3] {Michael A. Tabak}
\author[4]{Carl Chalmers}
\author[4] {Paul Fergus}
\author[1]{John Fieberg}
\affil[1]{Department of Fisheries, Wildlife and Conservation Biology, University of Minnesota, Saint Paul, MN, USA.}
\affil[2]{Departamento de Ciencias Biol{\'o}gicas, Universidad de los Andes, Bogot{\'a}, Colombia}
\affil[3]{Western EcoSystems Technology, ULC; 1000 9th Ave SW Suite 303; Calgary, AB T2P 2Y6}
\affil[4]{School of Computer Science and Mathematics, Liverpool John Moores University, Byrom Street, Liverpool, L3 3AF, UK}

\date{\today}

\begin{document}

\maketitle

\section{Abstract}

\begin{enumerate}
\item Camera traps have quickly transformed the way in which many ecologists study the distribution of wildlife species, their activity patterns, and interactions among members of the same ecological community. Although they provide a cost-effective method for monitoring multiple species over large spatial and temporal scales, the time required to process the data can limit the efficiency of camera-trap surveys.  Thus, there has been considerable attention given to the use of Artificial Intelligence (AI), specifically Deep Learning (DL), to help process camera-trap data. Using DL for these applications involves training algorithms, such as Convolutional Neural Networks (CNNs), to use particular features in the camera-trap images to automatically detect objects (e.g., animals, humans, vehicles) and to classify any species that are present. 
\item	To help overcome the technical challenges associated with training CNNs, several research communities have recently developed platforms that incorporate DL in easy-to-use interfaces. We review key characteristics of four AI-powered platforms -- Wildlife Insights (WI), Machine Learning for Wildlife Image Classification (MLWIC2), MegaDetector (MD), and Conservation AI -- including their software and programming requirements, data management tools, and AI features. We also provide R code and data from our own work to demonstrate how users can evaluate model performance using common  metrics (e.g., precision, recall, F1 score), and we discuss how these platforms can be used in conjunction with semi-automated workflows. 
\item	We found that species classifications from WI and MLWIC2 generally had low recall values (animals that were present in the images often were not classified to the correct species).  Yet, the precision of WI and MLWIC2 classifications for some species was high (when classifications were made, they were generally accurate). MD, which classifies images using broader categories (e.g., "blank" or "animal"), also performed well. Thus, we conclude that although species classifiers were not accurate enough to automate image processing, DL could be used to improve efficiencies by accepting classifications with high confidence values for certain species or by filtering images containing blanks. 
\item	By reviewing features of popular AI-enabled platforms and sharing examples via an open-source GitBook, we hope to facilitate the use of AI by ecologists to process their camera-trap data.
\end{enumerate}

Keywords: camera traps, artificial intelligence, deep learning, data processing, Wildlife Insights, MegaDetector, Machine Learning for Wildlife Image Classification 2, Conservation AI

\newpage
\section*{Introduction}

Camera traps are frequently used in ecological research to study animal behavior and to estimate  density, relative abundance, or occupancy in single- and  multiple-species studies \citep{burton_2015}. Camera traps can generate tremendous amounts of image data, and thus, much attention has been given recently to developing automated approaches for processing photos using Deep Learning (DL) algorithms. These algorithms can perform image classification and object detection after being trained using a pre-labelled dataset that uniquely identifies each species (or category) of interest.  DL has been widely used for removing photos that are "blank" (i.e., photos without animals) \citep{beery_2018}, species identification \citep{carl_2020}, individual recognition \citep{bogucki_2018, chen_2020}, and counting of individuals \citep{norouzzadeh_2018}. Others have reviewed and compared the performance of different state-of-the-art classification methods and DL architectures for identifying species in camera-trap photos \citep{norouzzadeh_2018, schneider_etal_2018, tabak_2018} and videos \citep{chen_2019}. Although DL makes it possible to process millions of pictures in short time periods (e.g., 1.2 million images in 24h), large and diverse amounts of pre-processed data may be required to train models, and the performance of DL approaches may suffer when models are applied to new environments \citep{beery_2018, tabak_2018}.

Inherent challenges associated with automated processing of photos using DL have been widely discussed. These include poor performance when models are developed using unbalanced training data sets (e.g., with highly variable numbers of images of each species) \citep{gomez_2017}, using small and geographically limited data sets but then applying the model more broadly \citep{schneider_2020}, or using low-resolution images for model training. Additionally, model creation and refinement require technical and programming expertise beyond the limits of many ecologists \citep{tabak_2020, christin_2019}. For rare species, users may need to use specialized techniques to increase the size of the training data set. For example, image sets can be augmented with images generated by simulating animals on empty photos and modifying features such as animal pose, illumination, and orientation \citep{beery2020synthetic}.  It can also be useful to identify particular species or sites where models perform poorly, and then use data from those sites to further train available models.

To reach a wider audience of camera-trap users, several initiatives have recently been launched with the goal of training DL models with broad and diverse image data sets and creating platforms that facilitate the use of AI via simple user interfaces and software (e.g., Wildlife Insights (WI), MegaDetector (MD), Machine Learning for Wildlife Image Classification (MLWIC2), and Conservation AI). These platforms differ in several aspects including their ease of use, required computer and programming skills, data management tools, and whether they focus only on coarse categorization of images or include the ability to classify species. Thus, platforms may be more or less suitable, depending on the user's needs. 

In addition to providing access to trained DL models, AI-based platforms can enable users to record additional information when viewing photos, including specific animal features  (e.g., age, sex, stripe or spot patterns, etc) that can facilitate further analyses. For example, uniquely identifying characteristics  may allow estimation of species density or abundance using spatial capture-recapture methods \citep{augustine_spim, royle_scr, efford_secr}. Other specific animal features, such as animal health characteristics, group sizes, or animal behavior \citep{norouzzadeh_2018} might also be of interest, as well as environmental conditions or signs of human activity within the camera's field of view \citep{greenberg_2019}.

\citeauthor{greenberg2020automated} (2020) discussed important aspects that need to be considered before using DL for automated image recognition, including knowing characteristics of the training data set (e.g., species included, number of pictures per species, and geographical locations of the image data). Additionally, he emphasized the need to use human verification to account for errors in DL output and provided a series of recommendations for processing camera-trap data using AI. Specifically, he recommended that users filter images with high confidence values associated with their AI classifications, and then review these images using bulk actions (e.g., selecting multiple species and accepting AI labels or correcting wrong labels provided by AI). In doing so, users can quickly accept classifications for categories that are likely to be correctly labeled (e.g., blank photos or species that are well represented in the training data set).

We build on this work by providing an overview of some of the AI-based platforms currently available to the public, along with possible workflows for processing camera-trap data. In section 2, we provide an overview of fully- and semi-automated image processing workflows. In section 3, we review different AI platforms and discuss their features for data upload, image identification, model training, and post-processing of classified photos. We consider platforms with diverse characteristics to illustrate a wide range of options; these platforms were also selected based on our perception of their stability and developer responsiveness. We also summarise pros and cons of these different platforms, thus providing readers with a quick reference or filter for choosing platforms that will fit their particular needs.  In section 4, we summarize results from a case study where we evaluate the performance of AI platforms for object detection and species classification using 112,247 photos collected from 50 camera traps deployed in the Colombian Orinoquia between January and July 2020.  Finally, we discuss the implications of our findings for choosing platforms and  workflows that incorporate AI for processing camera-trap data. We provide a more detailed overview of each AI platform and code for evaluating AI performance through the Data Repository of the University of Minnesota and an open-source GitBook \citep{gitbook_2022}.


\section{Workflows: fully- vs. semi-automated recognition}

Fully-automated recognition refers to pipelines in which \textit{computer vision} is used for detecting and identifying species or features in images without human review.  A fully-automated workflow is particularly useful for projects that require near-real-time detection of poachers or loggers, for preventing human-wildlife conflict, or for protecting species of high conservation concern that require immediate action in response to a threat. Other situations where a fully-automated recognition system might be useful include long-term projects with limited human capacity for image processing, projects with multiple deployments in the same geographical region, and projects that do not require further data annotation by humans to record various features in the image (e.g., environmental covariates and individual characteristics of the detected animals not considered in model training).

Although a fully-automated recognition workflow might sound appealing, it requires a trained model capable of providing highly accurate classifications. Users that desire a fully-automated workflow will likely need to train their own models using  data collected from their specific area of interest to ensure classifications are accurate. Users should also be aware that model performance may vary by species, and the impact of mis-classifications will depend on the underlying objectives, analysis approach, and target of estimation  \citep{whytock_2021}.

Most users will find that they need to implement a semi-automated workflow incorporating human review of classified images to meet their study objectives. AI platforms can accelerate image review by experts, and typically provide an image processing infrastructure that allows users to both verify DL model output (e.g., by accepting or rejecting model predictions) and to capture other characteristics of the images or the animals that are detected in each image.

Semi-automated workflows may accelerate image review by using AI output to filter and group images by categories that can be easily inspected \citep{greenberg2020automated}. For example,  empty photos (i.e., blanks) or photos containing particular species with high confidence values (i.e., species with a high probability of being correctly labeled by the model) can be filtered and quickly reviewed and verified using batch image selection. Some software packages, such as Timelapse 2 and Camelot, allow the user to interactively change the confidence value when selecting and filtering data; inspecting photos across a range of confidence values can help with determining an appropriate threshold for batch processing. Different confidence values may by appropriate for different species in the data set. Another common feature provided by many platforms, including MD (and its integration with Timelapse 2) and WI, is the display of bounding boxes around detected animals. Bounding boxes can be particularly useful for locating small mammals and birds.

\section{Overview of popular platforms}

\subsection{Which platform should I use?}

An initial determining factor in selecting an appropriate platform is whether users have data that can be made public.
Some platforms, such as MD and MLWIC2, were developed to maintain private workflows, while others, such as WI, are oriented towards open data and public data repositories. Additionally,  platforms differ in their ease of use, and users' operating system and internet access may also play a role in determining an appropriate platform. Another important consideration is whether users only need to discriminate between blanks and images with an animal or whether they need accurate species classifications. Because it can be difficult to achieve high accuracy rates when existing models are applied to novel data and environments \citep{schneider_2020}, users will typically want to select a platform that allows them to easily review images (using bulk selection/verification of images and image sorting/filtering) along with AI output so they can correct mis-classified photos. AI-powered platforms can also facilitate image inspection and handling (e.g., by providing a zoom feature that allows users to magnify or edit parts of an image), data entry and metadata extraction (e.g., date, time, temperature and lunar phase) \citep{greenberg_2019}.

\subsection{Wildlife Insights -- WI}

WI is an initiative developed by a partnership between Conservation International, Wildlife Conservation Society, World Wildlife Fund, Zoological Society of London, the Smithsonian Institution, North Carolina Museum of Natural Sciences, Yale University, and Google \citep{ahumada_wi}. WI serves as a data library and data-sharing platform in the cloud. Users can upload labeled or unlabeled images, through a Web-based upload tool, an application programming interface (API), or a desktop client. To promote data sharing and research collaboration addressing ecological questions at regional or global scales (e.g., assessment of species declines in response to climate change), WI  requires users to share their data under a Creative Commons license (CC0, CC BY 4.0, or CC BY-NC 4.0) after a maximum embargo period of 48 months. Other users can download data from the image repository using filters provided by the interface (e.g., to select for particular species, regions, dates). Public downloads will not contain exact coordinates of records of threatened terrestrial vertebrates (Critically Endangered (CR), Endangered (EN) or Vulnerable (VU) based on the IUCN Red List), to prevent exposure of geographical location of species that might be at risk (\href{https://www.wildlifeinsights.org/sensitive-species}{https://www.wildlifeinsights.org/sensitive-species, accessed on 01/13/2022}).

WI also provides tools for using AI to detect blank pictures and to identify over 800 different animal species from around the world (\href{https://www.wildlifeinsights.org/about-wildlife-insights-ai}{https://www.wildlifeinsights.org/about-wildlife-insights-ai, accessed on 01/13/2022}). WI uses a model trained using EfficientNet Convolutional Neural Networks for image classification with labeled camera-trap images collected by WI partners at sites worldwide (Table \ref{tab:train_source}). WI also provides bounding boxes in the interface, which is powered by a custom object-detection model. After uploading images to WI, they will be processed using AI, and then the user can download the resulting species classifications and metadata (e.g., time and date), which is automatically extracted by the system \citep{ahumada_wi}. Users can organize images hierarchically (e.g, by projects, sub-projects, and deployments), and therefore WI can serve as a useful project management tool. WI includes an interface to facilitate image processing and verification of AI output and to allow users to annotate images with additional information (e.g., specific animal features). Images can be processed in bursts (i.e., by grouping images within a time frame), and the cloud-based infrastructure makes it easy for multiple collaborators to process images simultaneously. WI also includes an analysis module that provides various data summaries, including species richness, species accumulation curves, and detection rates.

\textbf{Pros: } Includes tools for implementing efficient data processing workflows, managing collaborations (e.g., by  assigning different roles to group members), and sophisticated reporting and analytical capabilities. Serves as an image repository useful for data storage and data sharing for collaborative research. Users can edit AI predictions to improve future model performance.

\textbf{Cons: } Mandatory data sharing after an embargo period. Cloud-based, which makes it susceptible to connection instability and service outages (e.g., for system updates).

\subsection{MegaDetector}

Developed by the Microsoft AI for Earth program, MD is a model trained to detect blanks, animals, people, and vehicles from camera-trap images \citep{beery2019efficient}, and is also trained using data collected at a global scale (Table \ref{tab:train_source}). The model is based on the Faster-RCNN object-detection system with an InceptionResNetv2 base network and is hosted in the \href{https://github.com/microsoft/CameraTraps}{Microsoft/CameraTraps GitHub repository}, where it can be downloaded by users that want to run the model on their own \citep{beery2019efficient}. Around 10,000 images per day can be processed when using a standard computer and around 200,000 images can be processed per day when using a Graphics Processing Unit (which performs efficient computations by doing them in parallel). To run MD, users will need to be comfortable running computer code at the command line. Alternatively, users can contact the Microsoft AI for Earth camera-trap team who will then run MegaDetector for them once their data are uploaded to storage on Microsoft's cloud platform. AI for Earth will provide instructions for the data upload using a command-line utility for data transfer. Although data will need to be visible to Microsoft during processing in this scenario, they will not be shared or released publicly.

MD provides a JSON file as output, which indicates the locations of detected objects in each image and associated confidence values for each detection. Users can  run a Python script to sort, move, and organize pictures according to the MD predictions. When performing this task, users can choose a specific confidence threshold (CT) for determining which classifications should be accepted versus considered blank (e.g., using a 0.8 CT would classify images with confidence values less than 0.8  as "Blank"). MD can also facilitate an extra  post-processing step to reduce false positives (e.g., due to vegetation or background features that MD identifies as an animal), thereby increasing model accuracy. This step, implemented using a python script (see \href{https://github.com/microsoft/CameraTraps/tree/master/api/batch_processing/postprocessing/repeat_detection_elimination}{Microsoft/CameraTraps/api/batch/postprocessing/ GitHub repository}), involves identifying detections that have exactly the same bounding box across many images. Users can further process MD output using other platforms, such as Timelapse 2 and Camelot, as part of a semi-automated workflow to classify species.

\textbf{Pros: } Easy to integrate with other platforms (Timelapse 2, Camelot), provides different options for running the model (e.g., locally depending on user's computer capacity or with the help of the MD team). The object-detection framework allows the model to find multiple types of objects in the same image and facilitates object counting.

\textbf{Cons: } Detection of general categories (i.e., blanks, animals, people, and vehicles) instead of species labels.

\subsubsection{Timelapse 2}

Timelapse 2 is a software program for photo processing that can be run offline and in most versions of Microsoft Windows. Timelapse 2 incorporates AI results provided by the MD to accelerate further data processing. Timelapse 2 includes a Template Editor to allow the user to have complete control of any additional fields that they would like to record (e.g., vegetation characteristics associated with images, specific animal features). The Template Editor allows the user to specify a data-entry protocol, including the option of specifying data labels with default values and data-input controls that can prevent  errors when multiple people are involved in processing the images \citep{greenberg_2019}. 
To divide work between collaborators, images can be split by regions, locations, etc., and different MD results files can then be generated for each group of images. The images and the MD results files need to be transferred to collaborators using hard drives or cloud-based transfer and stored locally where Timelapse 2 will run. Once images and the MD results  are imported to Timelapse 2, users can start data processing and make use of the AI results to accelerate image revision. For example, users will be able to display all the images identified as blanks by computer vision with high confidence, allowing these images to be easily selected and marked as blanks.

\textbf{Pros: } Can incorporate MD output in data processing workflows, wide variety of processing options. No internet connection required after receiving the MD results file. Software stability.

\textbf{Cons: } Images need to be split and stored locally by each collaborator. Only runs on Windows computers.

\subsubsection{Camelot}

 Camelot was also developed for data management and processing purposes. It provides specific releases for Windows, OSX, and Linux operating systems, and a Java .jar release that can be used with any operating system. Users have to input camera deployment information, including a name and geographical coordinates associated with each camera and the dates it was in operation, either by providing the information via a Graphical User Interface or as a bulk CSV data import. Images can be imported to the software by browsing files on a local computer and will be presented in a \textit{Library} that serves as a dashboard where the user can visualize images, select one or multiple images at a time, edit their brightness and contrast, and inspect metadata associated with each image. Users have complete flexibility when specifying data fields to be recorded when processing data (e.g., specific animal features). 

Output from MD can be incorporated into Camelot to facilitate a semi-automated workflow, where users can filter images containing wildlife or people. This option requires a Camelot account and a good internet connection as it is an online service. After registering and uploading images to the cloud, users must activate the “wildlife detection” option in the “administration interface”, and Camelot will automatically run MD on these images. When activating image recognition using MD, users must provide an initial CT for assigning predictions made by computer vision, but this threshold can be changed at any time.

Camelot includes an analytical module that provides a summary of the percentage of nocturnal images and a Relative Abundance Index. It also generates summary tables that can be read into R using the camtrapR package for managing, visualizing, and tabulating camera-trap data \citep{camtrapR}. Camelot can also output detection matrices that can be used to fit occupancy models in Program PRESENCE \citep{hines_presence,mackenzie_occupancy}, and it allows users to thin data using a specified temporal independence threshold  \citep{iannarilli_lorelogram}. Camelot's web interface allows multiple users to work on the same project; the project owner can give remote access to collaborators by sharing the "Known URLs" found under the "Administration Interface" displayed when the application is opened. Camelot uses a Java virtual Machine to run and has minimum physical memory requirements of 2084 MB and 4096 MB for data sets of approximately 50,000 and 100,000 pictures. More details of memory limitations and options for working with large data sets can be found in the software documentation at \href{https://camelot-project.readthedocs.io/en/latest/}{https://camelot-project.readthedocs.io/en/latest/}.

\textbf{Pros: } Advanced reporting and analytical capabilities, can incorporate MD output in data processing workflow, wide variety of reports and processing options. Internet connection is not needed except when running AI models and when working with multiple collaborators.

\textbf{Cons: } Tasks (e.g., image upload, searching images, and summarizing output) can slow down as the data set increases in size. Users might need to manually configure Java for more efficient memory allocation when running Camelot.

\subsection{Machine Learning for Wildlife Image Classification -- MLWIC2}

MLWIC2 is an R package developed for detecting and classifying species from North America, although is also useful for identifying blank images using data collected at a global scale \citep{tabak_2020}. MLWIC2 allows the user to run its AI model on the user's device and to have an independent workflow without the need of image submission. Users need to install Anaconda Navigator, Python (3.5, 3.6 or 3.7), Rtools (for Windows computers), and version 1.14 of TensorFlow \citep{tensorflow2015-whitepaper} (GitHub repo, https://github.com/mikeyEcology/MLWIC2). Before running the model, users must pass the localization of Python in the \texttt{MLWIC2::setup} function. Users must know the R language and be familiar with file path specifications. MLWIC2 will provide an output file containing image filenames and the top five predictions for each image along with their associated confidence values.  Additionally, the R package provides functionality to train your own model using a subset of labeled images, which could be useful for improving AI performance. We illustrate the process used to train a model in the GitBook \citep{gitbook_2022} using a small set of images since training a model can be computationally intensive.

\textbf{Pros: } Models can be run locally once the package and associated tools are correctly installed. Provides a module for training your own model. Has a Shiny App for interactively using its AI model, and training your own model.

\textbf{Cons: } Requires more advanced computational skills and local computing power. Trained models are geographically limited to species from North America. As it requires past versions of Python and TensorFlow, the installation can be cumbersome.  MLWIC2 operation can be inconsistent between different computers due to using R to interface with Python.

\subsection{Conservation AI}

Conservation AI is a cloud-based platform developed at the Liverpool John Moores University (UK) to help conservation projects use AI to process acoustic recordings, drone images, and camera-trap pictures and videos. It currently has trained models for identifying humans, man-made objects (e.g., cars and fires), and species from the United Kingdom, South Africa, North America, and Tanzania. It provides services for image detection and classification in near-real time from linked devices capable of transferring images using a Simple Mail Transfer Protocol (SMTP). Any camera can be used for real-time detection as long as it supports SMTP and you have internet coverage in your study area. Alternatively, images can be directly uploaded to the platform using a batch upload of up to 1,000 pictures at a time. Once uploaded, images can be classified using the available AI models, which can process approximately 50,000 images per hour. Once images are uploaded and classified, the results will be available in the platform.

In addition to the currently available models, Conservation AI also provides a platform for image tagging and model training for specific data sets. Users can upload pictures directly into the tagging site or share them with the developers (e.g., via Google Drive) who will then upload batches of 500 pictures for you. For tagging, users will draw bounding boxes around animals in the images and label them with the species' name; this process will create the training data set. Users will need to tag a minimum of 1,000 images per species, and the available models will be updated using transfer learning based on the new tags. The tagging section contains a species list with tags from different projects registered in the platform, and users can request to train models using any of the tagged data available in the platform. Conservation AI provides all of its functionality in the cloud, so a good internet connection is needed. This platform will output species identifications along with associated confidence values for each record. 

\textbf{Pros:} Real-time detection capabilities, provides an easy-to-use platform for image tagging and model training.

\textbf{Cons:} Cloud-based, which makes it susceptible to connection instability and service outages (e.g., for system updates). When tagging species for model training, users must request Conservation AI developers to upload batches of 500 photos to the tagging site, which can take a few weeks depending on the developers' availability.

\section{Evaluating model performance}

Camera-trap users interested in incorporating AI into their workflows will need to evaluate model performance. This will require manually classifying a subset of images to species or to a broader set of classes (e.g., blank, human, animal). These \textit{human vision} labels can then be compared to \textit{computer vision} labels from an AI model to identify which, if any, species (or classes) are most likely to be correctly predicted. We illustrate this process in an open-source GitBook for WI, MD, and MLWIC2 \citep{gitbook_2022}. We did not evaluate the performance of Conservation AI because their available models for classification do not include species from South America. 

We evaluated model performance using data from a camera-trap survey conducted between January and July 2020 for wildlife detection within the private natural reserves El Rey Zamuro (31 km$^2$) and Las Unamas (40 km$^2$), located in the Meta department in the Orinoquia region in central Colombia. During the survey period, we collected 112,247 images from a 50-camera-trap array, with cameras spaced 1-km apart; 20 percent of the images were blank and 80 percent contained at least one animal. Images were stored and reviewed by experts using the WI platform. WI was chosen because it provides advanced processing capabilities that helped to accelerate image review (e.g., multiple image selection, image editing, and infrastructure for collaborative data processing). We release our images and annotations publicly in the Labeled Information Library of Alexandria: Biology and Conservation (\href{https://lila.science/orinoquia-camera-traps/}{https://lila.science/orinoquia-camera-traps/}), so they can be used as a benchmark training set that will encourage replication of our results and comparison with future AI tools. 

Expert (i.e., human vision) labels were compared to classifications by the AI models associated with WI (data downloaded in February 2021), MD (version 4.1), and MLWIC2 (version 1.0) to determine how well these models would perform when applied to data that were not included in their training data sets. Records containing the "Human" class were removed from the data set; these were predominately associated with images during camera setup. Workflows describing the use of the platforms, managing their output, and comparing predictions with labels from classified images using the R software \citep{R-base} are described in our online GitBook \citep{gitbook_2022}. Model performance was evaluated using functions in the \texttt{caret} package \citep{R-caret} in R to estimate a confusion matrix for the observed and predicted classes as well as precision, recall, and F1 scores (Table \ref{tab:metrics_def}). 

\section{Results}

\begin{figure}[ht]
     \centering
         \includegraphics[width=0.6\textwidth]{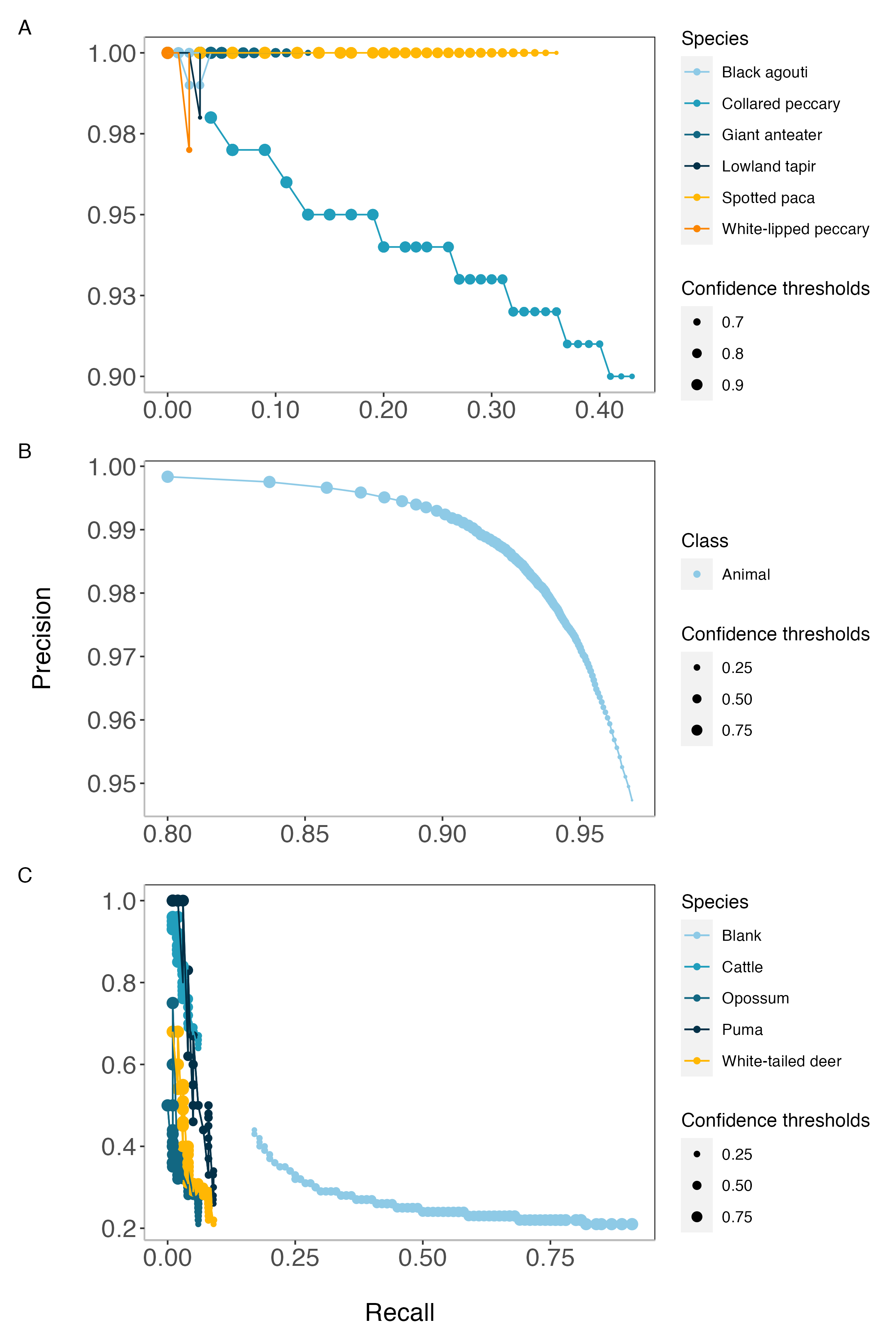}
              \caption{Precision and recall values for different confidence thresholds used to predict species or class labels using Wildlife Insights (A), MegaDetector (B) and MLWIC2 (C).}
              \label{fig:all_th}
     \end{figure}

Model performance varied widely between species and AI platforms (Table \ref{tab:metrics}). WI and MLWIC2 had high precision values for some species (at a CT = 0.65), suggesting that when computer vision predicted a species label, it was usually correct. However, all species had low recall values (less than 54\%, at a CT = 0.65), indicating that these platforms missed many of the animals present in the images (Table \ref{tab:metrics}). Species classifications for MLWIC2 tended to have low recall, likely due to strong differences between the training and test data sets. For this particular data set, WI's AI would be most useful for classifying collared peccaries and spotted pacas (both abundant species in the data set, representing 22\% and 5\% of the images with animal records, and with conspicuous fur patterns). We can be confident that WI's AI is correctly labeling collared peccaries and spotted pacas (precision of 90\% and 100\%, respectively at a CT = 0.65). Yet, it is only finding 43\% and 36\% of the records for these species (at a CT = 0.65) (Table \ref{tab:metrics}). These results highlight that AI platforms may be able to speed up the process of species classification by allowing users to accept classifications for species that have high precision. Yet, low recall values will necessitate expert review of photos labeled as "Blank" or containing species that have poor precision values.  I.e., these platforms would be best used in a semi-automated workflow where experts still review computer vision output.  Users can further increase precision by selecting images with high confidence values but at the cost of decreasing recall (Figure \ref{fig:all_th}).

\begin{figure}[ht]
     \centering
         \includegraphics[width=15cm, height=15cm]{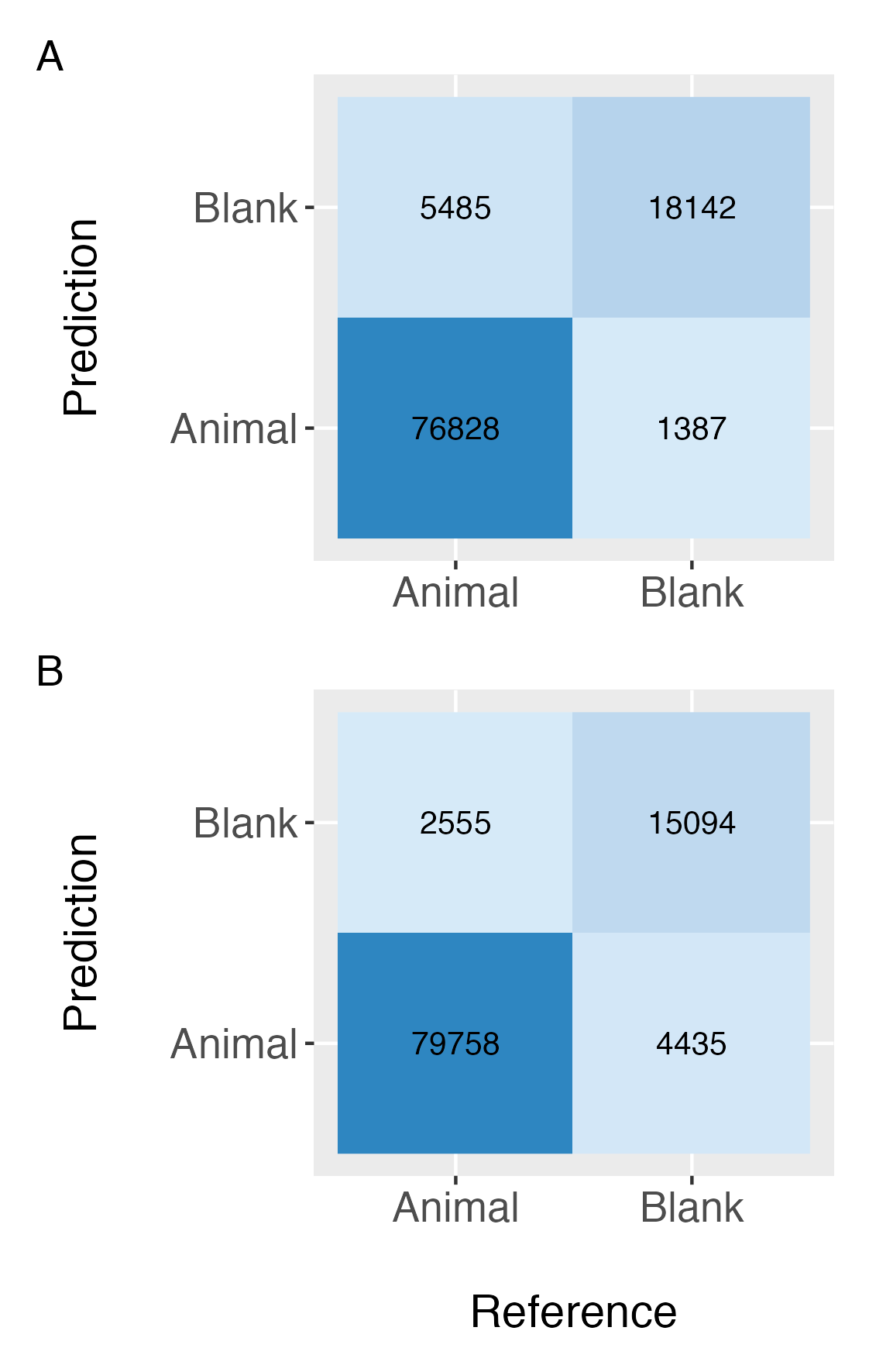}
              \caption{Confusion matrices comparing expert labels (Reference) and MegaDetector predictions using 0.65 (A) and 0.1 (B) as confidence thresholds to assign labels provided by computer vision.}
              \label{fig:cm_md}
     \end{figure}

MD, which identifies broader categories of objects, had a precision of 98\% at a 93\% recall (at a CT = 0.65) for the "Animal" class, and consequently also had a high F1 score (Table \ref{tab:metrics}). Thus, MD is extremely good at detecting animals in images, with very low probabilities of both false positives and false negatives (Table \ref{tab:metrics}). Decreasing the confidence threshold from 0.65 to 0.1 (Figure \ref{fig:cm_md})  increased the recall for the "Animal" class from 93\% to 97\%, but decreased precision from  98\% to 95\%. Thus, using a lower confidence threshold could reduce the number of false negatives associated with the "Animal" class. When using MD, users will still need to review images classified as having animals (to classify to the species level). Thus, it may be best to use lower confidence thresholds to maximize recall (so that animals are not missed) at the expense of having slightly lower precision.  

\section{Discussion}

Common challenges associated with image recognition using DL, such as low accuracy when classifying species at new locations \citep{schneider_2020}, and variable model performance for different species \citep{whytock_2021}, are persistent even when using models trained with broad and diverse image data sets. Despite these challenges, DL can help ecologists establish more efficient workflows for processing camera-trap images by providing accurate classifications for some species or by identifying blank images. Some AI platforms also provide additional functionality for managing and annotating large camera-trap data sets.  For example,  WI provides a comprehensive infrastructure for sharing, managing and storing camera-trap photos, and Timelapse 2 provides several useful tools for classifying and annotating images.

Although we found that AI platforms were able to  classify some species with high levels of precision, recall values were typically low; thus, experts will still need to review images to find the animals missed by computer vision. MD was useful for removing blank photos. Once blanks are removed, images can be integrated with other systems, such as WI, Timelapse 2 or Camelot, for further species classification or annotation by humans. Users interested in developing a fully-automated workflow for species classification will likely need to train their own models, for example, using the MLWIC2 package in R or Conservation AI's infrastructure.

The development of AI models for species identification is an area of active research, and the platforms we have reviewed are undergoing continuous model development. AI models continue to be updated with new data and should lead to better model performance over time. However, at least in the near term, most users will need to review AI classifications \citep{whytock_2021}, both to correct incorrect classifications and to capture other relevant information in the images (e.g., specific animal features). To date, there has been little work to develop AI models that can identify individual characteristics  (e.g., an animal's sex or age class) or behaviors (e.g., whether animals are feeding, moving, or resting). We expect DL will also play a significant role in predicting these characteristics and behaviors once more data have been collected and made available for training new models. 

\section{Acknowledgements}

We thank Juan David Rodríguez and volunteers for their assistance with camera trap data collection. We also appreciate the constant support from César Barrera and Eduardo Enciso that allowed us to set up camera traps on their properties and have facilitated field expeditions. We acknowledge the valuable comments on the manuscript provided by Tanya Birch from Wildlife Insights and the thorough review provided by Dan Morris from the Microsoft AI for Earth Program that considerably improved the manuscript. This research was made possible thanks to funding from the Colciencias - Fulbright Scholarship, the WWF’s Russell E. Train Education for Nature Program (EFN), the Interdisciplinary Center for the Study of Global Change Fellowship and the Department of Fisheries, Wildlife and Conservation Biology at the University of Minnesota. JF received partial salary support from the Minnesota Agricultural Experimental Station.

\section{Authorship}

JV and JF conceived the ideas and designed methodology; PC and JV collected and processed the data; JV and JF analysed the data and led the writing of the manuscript. All authors contributed critically to the drafts and gave final approval for publication.

\section{Conflicts of Interest}

We have no conflicts of interest.

\section{Data Archive}

Camera trap images and annotations are archived in the Labeled Information Library of Alexandria: Biology and Conservation (LILA BC) \href{https://lila.science/orinoquia-camera-traps/}{https://lila.science/orinoquia-camera-traps/}. Code and guidelines to use AI platforms for processing camera trap data are published in an open-source GitBook \citep{gitbook_2022}.

\newpage

\renewcommand\refname{References}

\bibliography{ai_bibliography.bib}

\newpage

\section*{Figures}

\newpage\clearpage

\newpage\clearpage

\section*{Tables}

\begin{table}[ht]
\centering
 \includegraphics[width=0.8\textwidth, center]{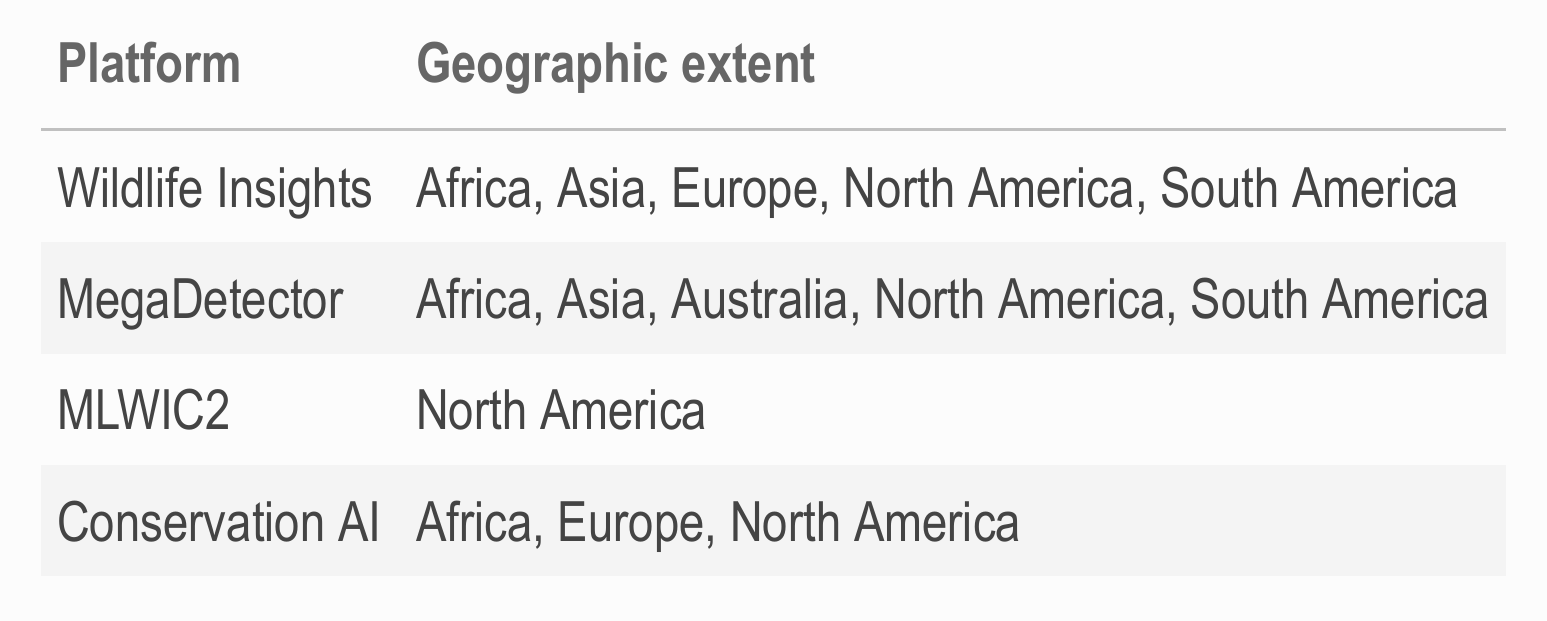}
    \caption{Geographical extent of training data used by Wildlife Insights, MegaDetector, MLWIC2 and Conservation AI.}
    \label{tab:train_source}
\end{table}

\newpage

\begin{table}[ht]
\centering
 \includegraphics[width=1\textwidth, center]{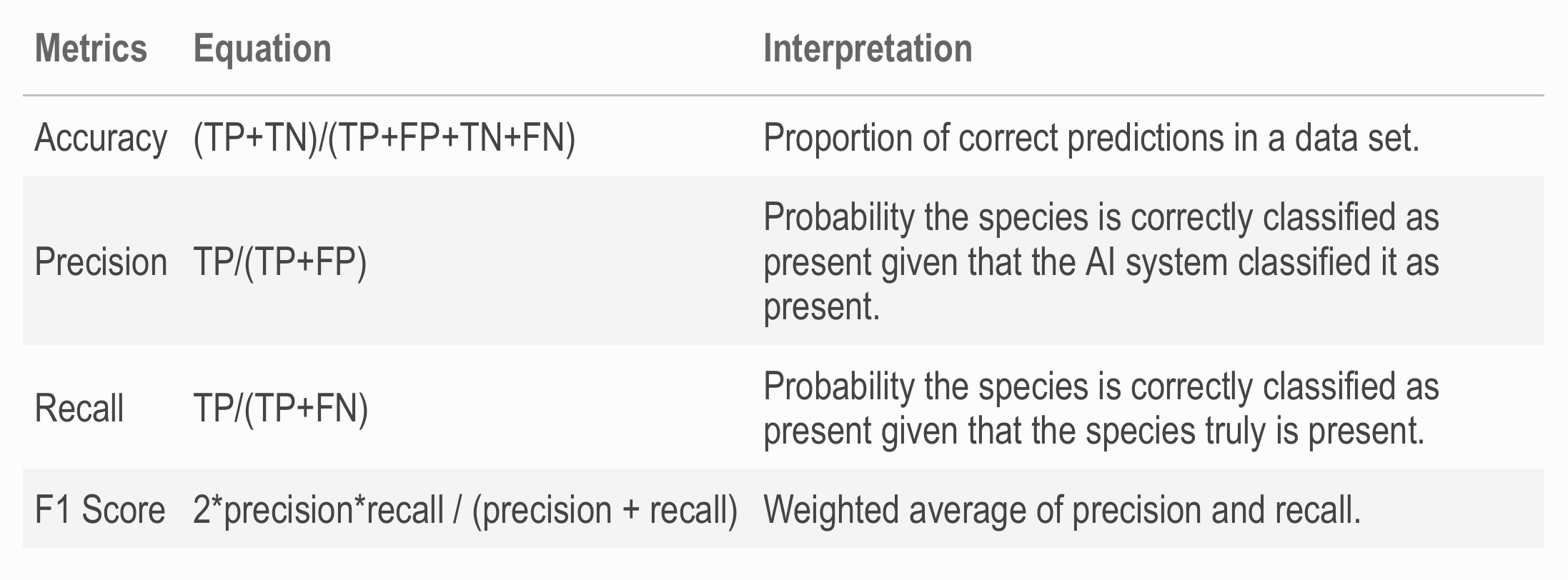}
    \caption{Metrics used to assess model performance. True positives (TP): Number of observations where the species was correctly identified as being present in the photo; True Negatives (TN): Number of observations where the species was correctly identified as being absent in the photo; False positives (FP): Number of observations where the species was absent, but the AI classified the species as being present; False negatives (FN): Number of observations where the species was present, but the AI classified the species as being absent.}
    \label{tab:metrics_def}
\end{table}

\begin{table}[ht]
\centering
 \includegraphics[width=2\textwidth, center]{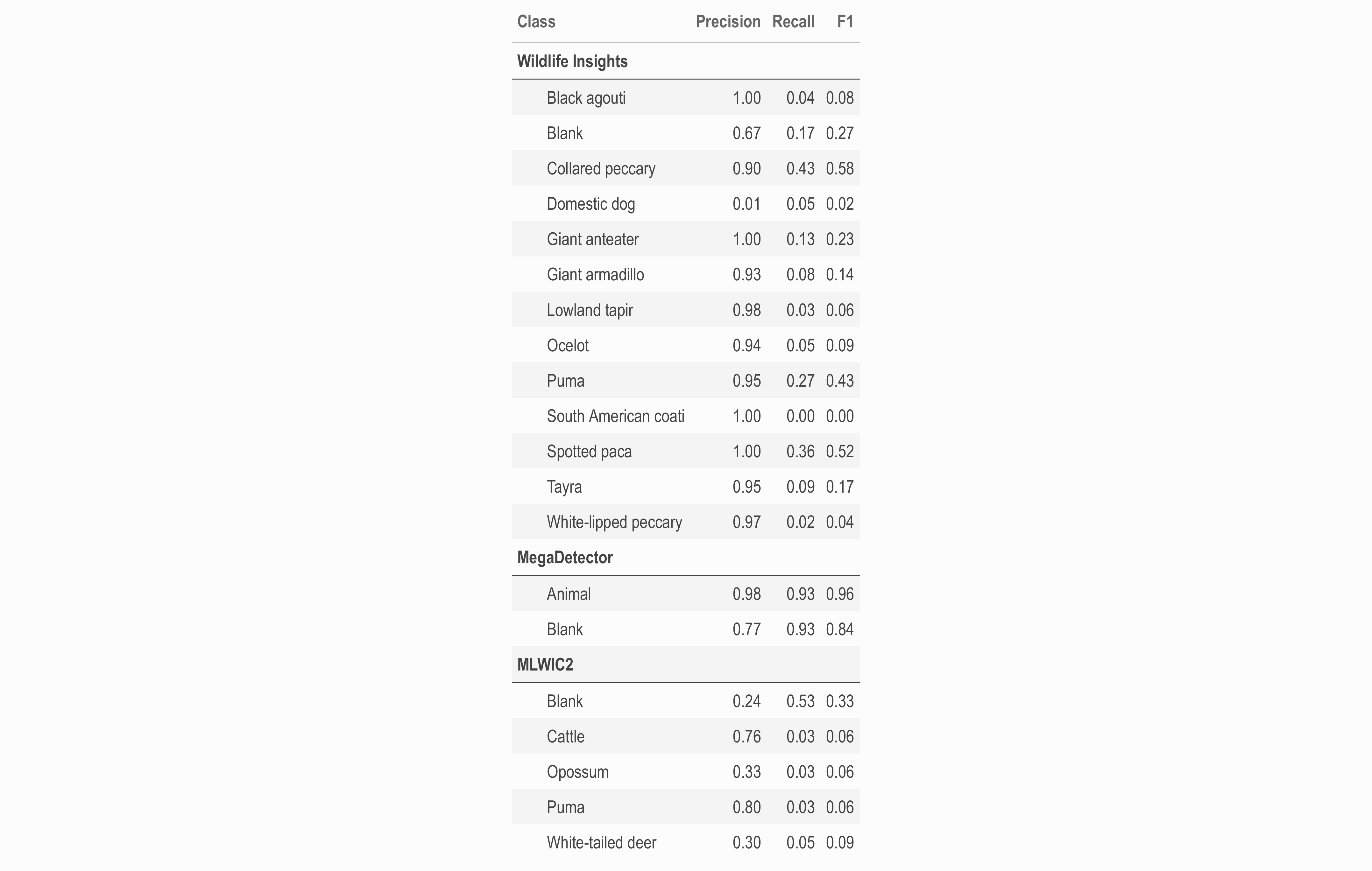}
    \caption{Model performance metrics for classes predicted by Wildlife Insights, MegaDetector and MLWIC2 using a confidence threshold of 0.65.}
    \label{tab:metrics}
\end{table}

\newpage\clearpage

\end{document}